# A Surrogate Model for the Forward Design of Multi-layered Metasurface-based Radar Absorbing Structures

Vineetha Joy[1*], Aditya Anand[2], Nidhi[1], Anshuman Kumar[3], Amit Sethi[3], Hema Singh[1]
[1]Centre for Electromagnetics, CSIR-National Aerospace Laboratories, Kodihalli, Bangalore 560017
[2]Birla Institute of Technology and Science, Pilani, Rajasthan 333031
[3]Indian Institute of Technology, Bombay, Maharashtra 400076
vineethajoy@nal.res.in, rgaditya@gmail.com; nidhiquantum1007@gmail.com; anshuman.kumar@iitb.ac.in, asethi@iitb.ac.in, hemasingh@nal.res.in

*Abstract*— **Metasurface-based radar absorbing structures (RAS) are highly preferred for applications like stealth technology, electromagnetic (EM) shielding, etc. due to their capability to achieve frequency selective absorption characteristics with minimal thickness and reduced weight penalty. However, the conventional approach for the EM design and optimization of these structures relies on forward simulations, using full wave simulation tools, to predict the electromagnetic (EM) response of candidate meta atoms. This process is computationally intensive, extremely time consuming and requires exploration of large design spaces. To overcome this challenge, we propose a surrogate model that significantly accelerates the prediction of EM responses of multi-layered metasurface-based RAS. A convolutional neural network (CNN) based architecture with Huber loss function has been employed to estimate the reflection characteristics of the RAS model. The proposed model achieved a cosine similarity of 99.9% and a mean square error of 0.001 within 1000 epochs of training. The efficiency of the model has been established via full wave simulations as well as experiment where it demonstrated significant reduction in computational time while maintaining high predictive accuracy.**

## I. Introduction

Metasurfaces are ultra-thin two dimensional structures capable of controlling the amplitude, phase and polarization of incident electromagnetic (EM) waves with added benefits like compactness and ease of fabrication [1]. Due to these attributes, they are widely being used for different applications like stealth technology [2], [3], [4], [5], electromagnetic shielding [6], [7], [8], wireless communication systems [9], [10], [11], holographic imaging [12], [13], etc. With respect to stealth technology, metasurface based radar absorbing structures (RAS) are used to reduce the radar cross-section (RCS) of hotspots on low observable platforms.

The EM design of metasurface based RAS for desired frequency selective reflection characteristics is a clear case of inverse design. The conventional approach typically involves brute-force optimization techniques like genetic algorithms, particle swarm optimization (PSO), etc. These algorithms rely on iteratively evaluating the performance of candidate metasurface configurations until a specific criterion is met. A critical component of this inverse design process is the accurate prediction of the EM response of individual meta-atoms based on their geometric and material parameters—commonly referred to as forward design. Such an integrated system that combines both the forward and inverse design processes in a complementary way is called tandem architecture [2]. The schematic of the tandem neural network model *w.r.t.* the present scenario is shown in Fig. 1.

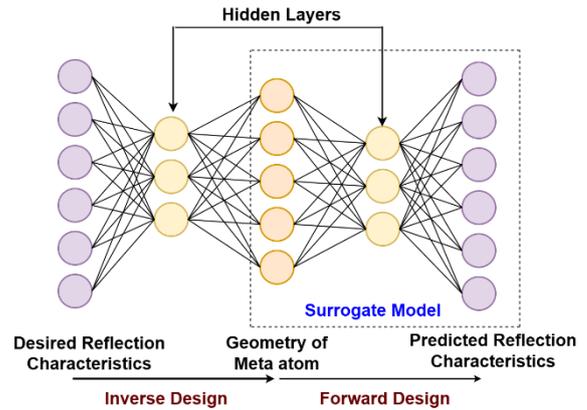

Fig. 1. Tandem architecture for design of metasurface based RAS.

Commercially available simulation tools based on full wave techniques like finite element method (FEM), finite-difference time-domain (FDTD) method, method of moments (MoM), etc. are generally used for the forward simulation. Although these approaches are reliable, they suffer from significant computational overhead, particularly when exploring large design spaces. These challenges have motivated the development of surrogate models which are fast, data-driven approximations that emulate the behavior of full wave EM simulators with significantly reduced computational cost.

In this regard, the present paper proposes a surrogate model for the forward design of multi-layered metasurface based RAS. Given the 2D image of the meta atom pattern along with the information on its configuration, the surrogate model predicts the reflection characteristics of the corresponding planar RAS. The proposed model maps geometric, material and configuration parameters of metasurface unit cells to their corresponding EM responses. A CNN based architecture with Huber loss function has been used to estimate the reflection characteristics of the RAS. Trained on a representative dataset of different classes of unit cell geometries, generated via full-wave simulations, the surrogate achieves high prediction accuracy and generalizes well across diverse design parameters. The performance of the model has been assessed using several examples across the broad frequency range of 2GHz – 18GHz. The predictive accuracy of the model is established via experiment as well.

## II. BACKGROUND AND RELATED WORK

Surrogate models approximate the complex input–output mappings between structural/material parameters of meta atoms and their corresponding EM responses with significantly reduced computational cost. Unlike physics based solvers like FDTD, FEM, MoM, etc, surrogate models rely on data-driven approaches to emulate the system behavior. In the context of metasurface design, surrogate models are particularly valuable due to the high dimensionality of the design space and the computational burden associated with simulating multi-layered structures. By learning from a curated dataset of simulated unit cells, these models can generalize to unseen geometries and support fast forward predictions. This attribute makes them especially suitable for inverse design and iterative optimization tasks.

Surrogate modeling techniques for metasurfaces or in general microwave structures can be categorized into two main types: physical and functional surrogates [15]. Commonly used physical surrogates are based on analytical and equivalent circuit models. Physical surrogates are built upon simplified models that retain a physical relationship to the original structure, offering faster evaluations. Although they are computationally inexpensive, they often suffer from limited accuracy and are generally unavailable for complex structures. Functional surrogates rely on data sampled from full wave EM simulations and use mathematical approximation methods to emulate the behavior of the structure. Functional surrogate models can be built using a wide range of function approximation methods such as low-order polynomial expansions, radial basis functions, Kriging, regression models, and neural networks. The prominent functional surrogate approaches are described below:

a) *Artificial neural networks (ANNs)*: This method stands out as the most widely adopted among functional surrogate modeling techniques. Fully connected deep neural networks (FCNN) have been used [16], [17] to predict the EM response of a single pre-identified meta atom (the periodically repeating structure in meta-surfaces) using its structural parameters in the form of an input vector. Similar architectures have been used [18], [19] to simulate the chiroptical response and circular dichroism response of chiral metamaterials. Another category of NNs namely physics-informed neural networks (PINNs) are designed to solve supervised learning tasks while incorporating physical laws, typically expressed as general nonlinear partial differential equations [20], [21], [22]. In the context of metasurfaces, PINNs incorporate Maxwell's equations and boundary conditions into the loss function thereby enhancing the prediction accuracy [23], [24], [25]. Further, in order to predict the EM response of limited classes of meta atoms, given the image of the pattern, convolutional neural network (CNN) based architectures have been used [26], [27] [28].

d) *Statistical or probabilistic models:* These models are widely used as a surrogate modeling technique in EM simulations due to its ability to provide accurate predictions along with uncertainty quantification [29]. Several research groups [30], [31] have demonstrated the utility of gaussian process regression (GPR) in the optimization of metasurfaces and microwave structures, especially under geometrical uncertainty, enabling data-efficient design in high-dimensional EM problems. Further, Kriging models have been used in the optimization of reflector antennas [32].

Despite growing interest in surrogate modelling for the design of metasurfaces, most of the approaches focus on single-layer configurations and narrow design spaces, limiting their applicability to complex practical scenarios. Majority of the surrogate models reported in literature can predict the EM response of only very few classes of meta atom geometries with many of the configuration parameters fixed including number of layers and materials. Further, the loss functions used in most of the models are not designed or chosen to handle the effect of outliers in EM response. In this direction, this work presents a novel surrogate model specifically tailored for the forward prediction of reflection characteristics of multi-layered metasurface-based radar absorbing structures. The main contributions of this paper are summarized as follows:

(i) Novel architecture: The proposed surrogate model features a carefully designed CNN architecture optimized specifically for learning the complex EM characteristics of multi-layered metasurface-based RAS. Extensive hyperparameter tuning and optimization were carried out to enhance model generalization and prediction accuracy.

(ii) Robust training via Huber loss: The surrogate model employs the Huber loss function during training, offering improved robustness against outliers commonly encountered in simulation datasets. This leads to more stable and reliable predictions.

(iii) Comprehensive coverage of meta atom geometries: The model is trained on a diverse and extensive dataset comprising of 16 distinct classes of meta-atoms covering different areas in the broad frequency range of 2GHz -18GHz. This comprehensive coverage enables the model to generalize across a wide spectrum of design geometries and structural variations.

(iv) Inclusion of multi-layered configurations: Unlike prior studies that primarily address single-layer metasurfaces, this work includes multi-layered designs as well, capturing inter-layer coupling effects that are important for accurate modelling of broadband absorption.

(v) Experimental validation: To ensure practical relevance, the proposed surrogate model is validated against experimentally measured absorption characteristics of fabricated RAS prototypes. This validation demonstrates the model's ability to accurately predict real-world electromagnetic behavior.

## III. ARCHITECTURE OF SURROGATE MODEL

The proposed surrogate model is composed of four convolutional blocks (Fig. 2a) followed by two levels of max pooling and three fully connected layers. Each convolutional block comprises of two successive convolutional layers, each followed by batch normalization and a LeakyReLU activation function. The block concludes with a max pooling layer to reduce spatial dimensionality and capture dominant features. The input to the model is the 2D gray scale image (1×500×500) of the pattern of meta atom. The output of the model is the predicted reflection spectra (1×201). In the model, the output of the last pooling layer is flattened and the resulting vector is concatenated with the configuration data corresponding to the meta atom. The configuration data includes details on material properties (permittivity, electric loss tangent, permeability, magnetic loss tangent, thickness), number of layers (single-layered or double-layered) and pattern material (metallic or resistive), etc. The complete architecture of the surrogate model is shown in Fig. 2b.

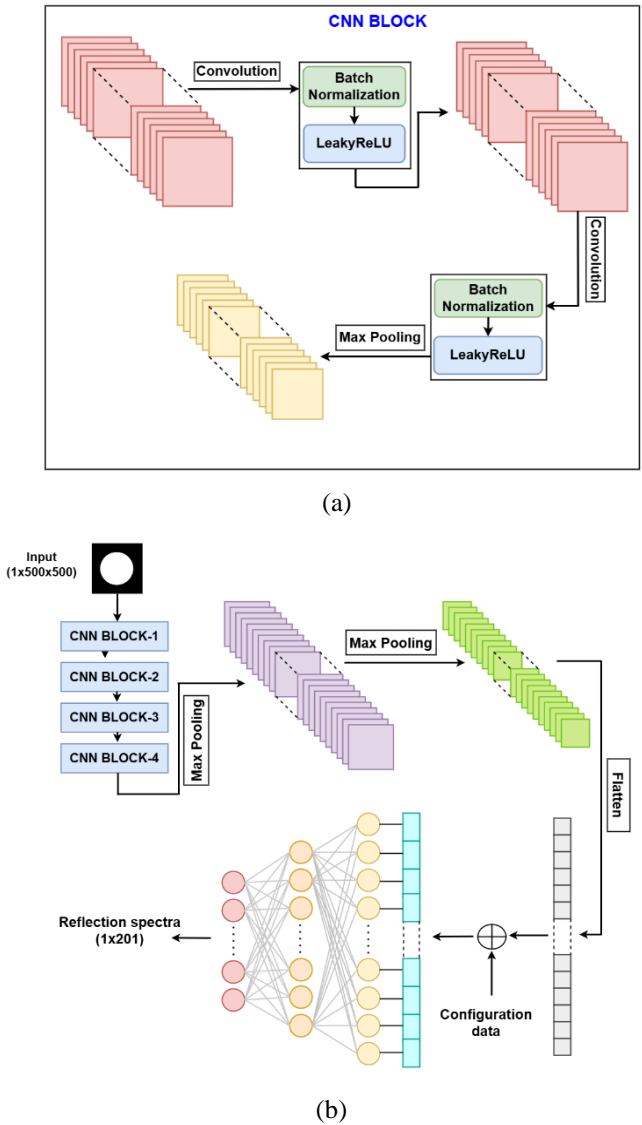

(a)

(b)

Fig. 2. Surrogate model for the forward design of metasurfaces. (a) Schematic of a CNN block (a) Complete architecture of the surrogate model.

The surrogate model has been trained using Adam optimizer to minimize the difference between the desired spectra and the predicted spectra. Here, Huber loss function has been used as it reduces the influence of outliers without completely ignoring them [33]. The loss function is really crucial in the present case, since the dataset used consists of reflection spectra with various spikes and dips, which are difficult to model with only mean square error (MSE) loss.

Huber loss is a combination of mean absolute error (MAE) and MSE. For small errors, it behaves like MSE, and for larger errors, it behaves like MAE, which is robust to outliers. The loss function can be expressed as,

$$\text{Huber Loss} = \frac{1}{N}\sum_{i=1}^{N}\begin{cases} \frac{1}{2}(y_i - \hat{y}_i)^2, & |y_i - \hat{y}_i| \leq \delta \\ \delta|y_i - \hat{y}_i| - \frac{1}{2}\delta^2, & |y_i - \hat{y}_i| > \delta \end{cases} \quad (1)$$

where $y^i$ and $\hat{y}^i$ are the actual and predicted values respectively. $N$ is the total number of samples. $\delta$ is a hyperparameter that determines the point at which the loss switches from MAE to MSE.

IV. PREPARATION OF DATASET

To ensure that the surrogate model achieves adequate generalization capability, the training dataset should include a diverse set of meta-atom geometries. These geometries should be capable of producing a wide range of reflection characteristics across different regions of the target frequency spectrum, encompassing both wideband and narrowband functionalities. In the present case, the surrogate model has been trained on a dataset of 16000 samples with over 16 classes of geometrically different meta atoms in both single-layered as well as dual-layered configurations. Each class has been simulated using 18 different commercially available substrate materials as well. These classes along with their appropriate parameter combinations have been carefully chosen to generate varieties of EM spectra in the frequency range of 2GHz-18GHz. The configuration of RAS models included in the dataset are shown in Fig. 3.

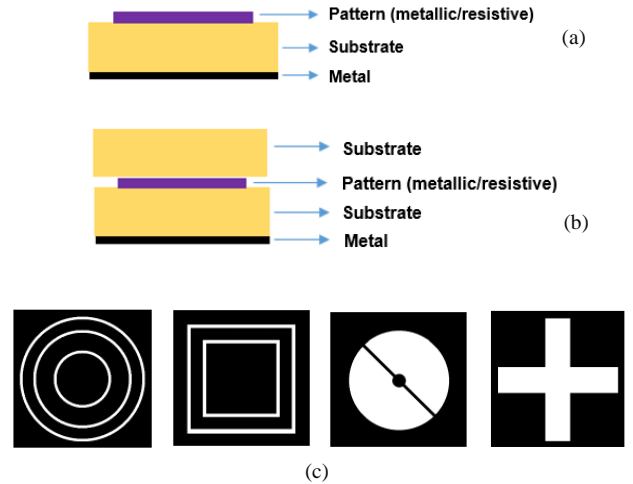

Fig. 3. Configuration of RAS models included in the dataset (a) Single-layered RAS (b) Dual-layered RAS (c) Samples of meta atom patterns [34].

The dataset has been generated via EM simulations in CST Studio Suite [35]. The meta atoms have been simulated using the frequency domain solver. Periodic boundary conditions were applied along the lateral directions to emulate an infinite array of unit cells. Along the direction normal to the surface of the meta atom, open (add space) boundary conditions were used. Plane wave excitation has been introduced to simulate the incident electromagnetic field. 99% of the data points have been used for training and 0.5% each has been used for validation and testing.

## V. RESULTS

$\delta$ is an important hyper parameter w.r.t Huber loss. In order to identify the optimal value of $\delta$, this hyperparameter has been varied from 0 to 3 and the corresponding variation in MSE and MAE during validation is shown in Fig. 4. It is clear from the plot that MSE is lowest for $\delta = 3$ and MAE is low as well for $\delta = 3$. Hence this value has been fixed for training of the model.

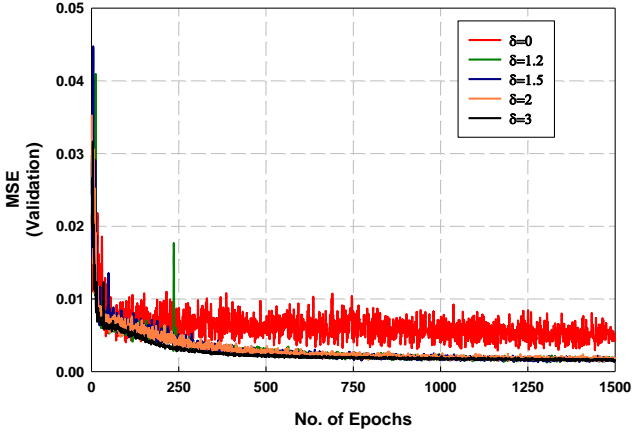

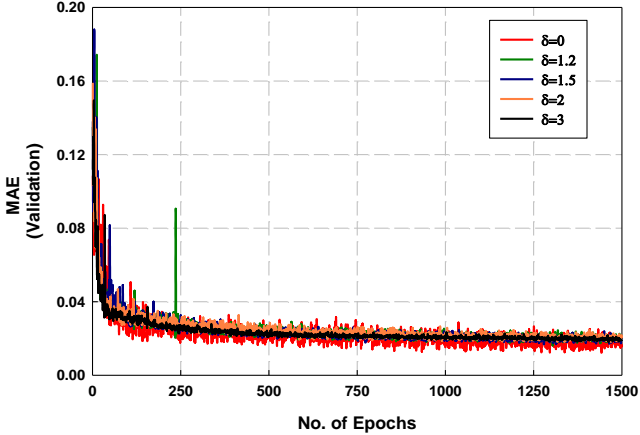

Fig. 4. Variation in losses w.r.t. hyperparameter, $\delta$ during validation (a) MSE (b) MAE.

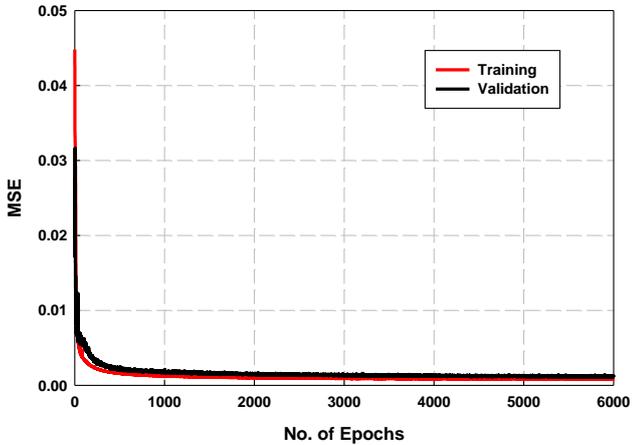

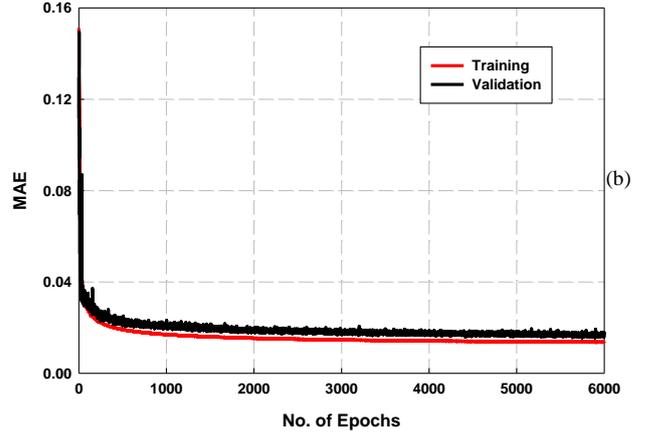

Fig. 5. Variation in losses of the model during training (a) MSE (b) MAE.

The model with the fixed set of hyperparameters (Learning rate = 0.0001; Optimizer: Adam ($\beta_1 = 0.5$; $\beta_2 = 0.999$); $\delta = 3$) has been trained for 6000 epochs. The variation in losses of the model *w.r.t.* number of epochs is shown in Fig. 5. The validation losses converged to very low values within very few epochs. The accuracy of the model has been evaluated using cosine similarity (CS) metric as well, which is a measure of the similarity between two non-zero vectors in an inner product space. CS between $\vec{A}$ and $\vec{B}$ can be calculated by the following expression:

$$CS = \frac{\vec{A} \cdot \vec{B}}{\|A\|\|B\|} \quad (2)$$

A cosine similarity of 1 means that the vectors are identical in direction. The variation in cosine similarity of the model *w.r.t.* number of epochs during training is shown in Fig. 6. The model achieved a cosine similarity of 99.8% on the validation data set within 6000 epochs of training.

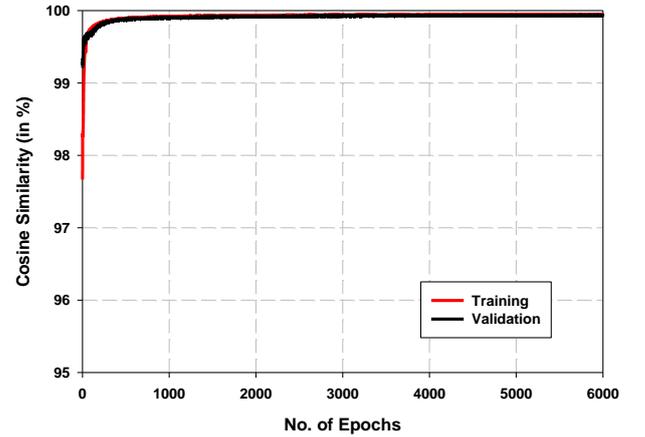

Fig. 6. Variation in CS of the model during training.

The predictive performance of the trained model has been tested by using the data points in the test data set. The 2D grayscale images of the meta atom patterns (shown as inset in Fig. 7) and the corresponding configuration details have been given as input to the trained model. The comparison between the desired spectra and the spectra predicted by the surrogate model for different data points are shown in Fig. 7.

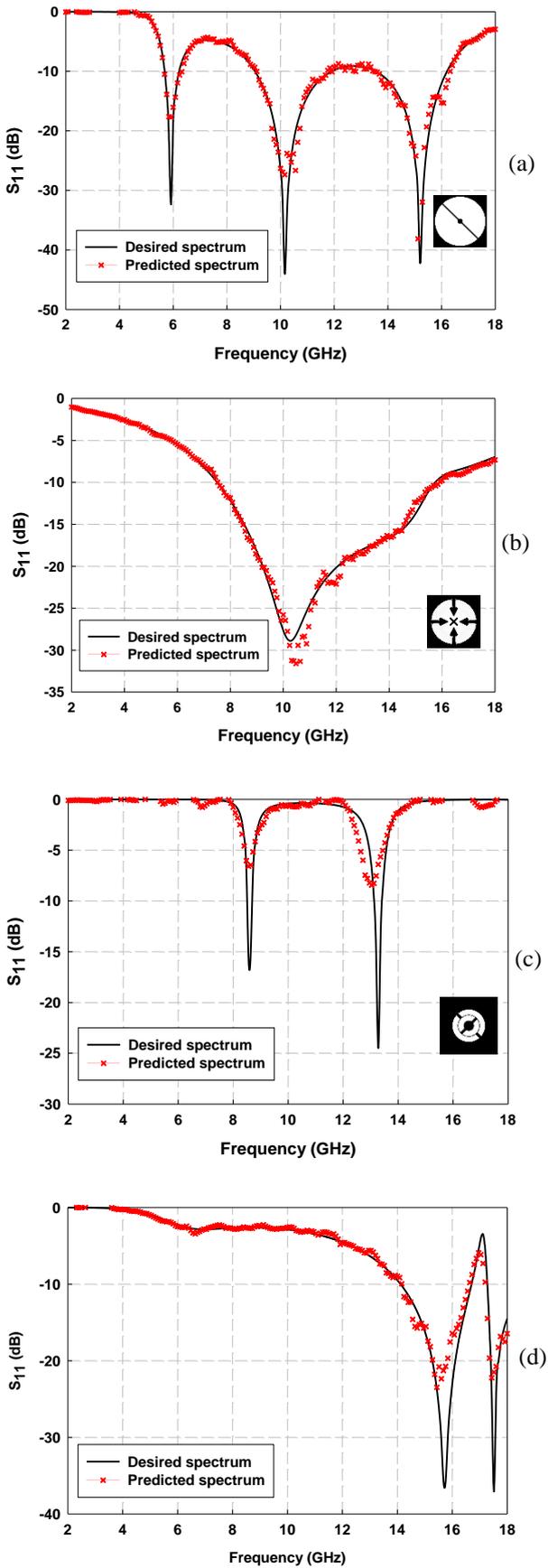

Fig. 7. Comparison between desired spectra and the spectra predicted the surrogate model for different samples in the validation set. (a) Sample-1 (b) Sample-2 (c) Sample-3 (d) Sample-4

The comparative plots clearly indicate that the desired spectra and the predicted spectra are in close agreement with each other. Outliers like reflection dips have been correctly predicted by the model establishing the efficacy of Huber loss. In the spectra predicted by the surrogate model, the frequency ranges in which magnitude of $S_{11}$ is less than -10dB has been found to be same as that in target spectra. It is also noted that the developed model took just six seconds for predicting the output whereas CST Studio Suite (commercially available EM simulation software) took 4 minutes for the same simulation using the same computational resources, i.e., there is 97.5% reduction in computational time.

## VI. EXPERIMENTAL DEMONSTRATION

In order to verify the predictive accuracy of the proposed model, the meta atom configuration as shown in Fig. 8a has been given as input to the trained surrogate model. The predicted output has been noted. A planar radar absorbing structure based on the meta atom has been fabricated and the prototype is shown in Fig. 8b. The variation in $S_{11}$ of the planar RAS at normal incidence for X-band has been measured using a vector network analyzer (model No. E8364B) and two spot focusing horn lens antenna (set up shown in Fig. 8c).

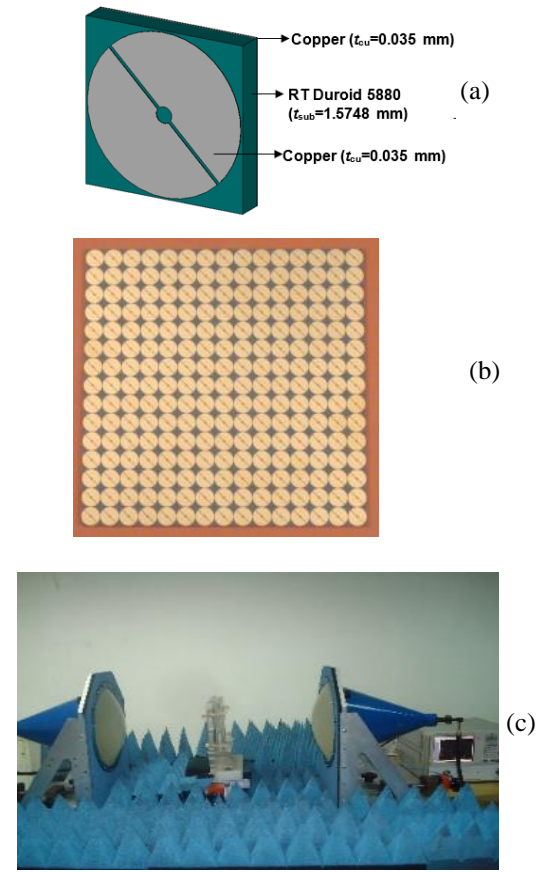

Fig. 8. Experimental demonstration (a) Configuration of meta atom (b) Fabricated RAS (c) Measurement set up.

The absorption characteristics have been then calculated using the following formula:

$$|A|^2 = (1 - S_{12}^2 - S_{11}^2) \qquad (2)$$

Here, $|A|^2$, $|S_{12}|^2$ and $|S_{11}|^2$ correspond to power absorbed, power transmitted and power reflected by the RAS respectively. $|S_{12}|^2$ will be zero in the present case as one side is completely covered by metal. The frequency dependent measured absorption characteristics of the fabricated RAS along with the predicted results are shown in Fig. 9. The plot clearly shows very good agreement between measured and predicted results establishing the reliability of the developed model for real-time applications. The measurement results also show that the power absorbed by the fabricated RAS is greater than 90% in the frequency range of 8.2 GHz to 12.4 GHz.

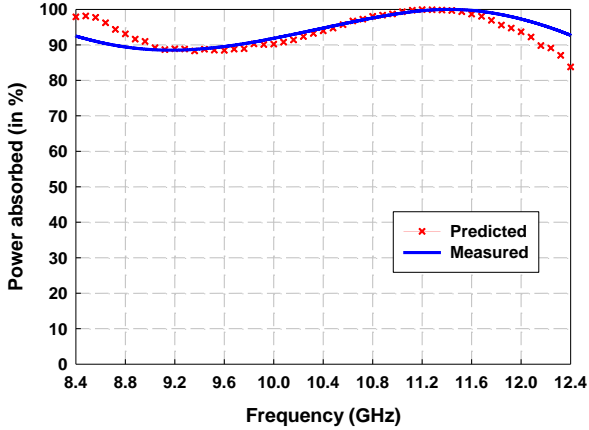

Fig. 9. Comparison between measured spectra and the spectra predicted the surrogate model.

## VII. CONCLUSION

In this paper, a surrogate model is presented for the computationally efficient forward design of multi-layered metasurface based RAS. By leveraging machine learning techniques, the proposed approach is shown to significantly reduce the computational cost associated with the conventional design approach while maintaining high accuracy. The model achieved significant reduction in simulation time in comparison with the time required by commercially available EM simulation software using the same computational resources. Comparative evaluations clearly show that the predicted spectra closely match the desired spectra. The cosine similarity of the model is around 99.9% which indicate high level of predictive accuracy. Further, the measurement results also show close agreement with the predicted results. The proposed surrogate model enables rapid exploration of design parameters, facilitating real-time optimization and inverse design. It can be used in combination with generative architectures like generative adversarial networks (GAN), diffusion models, etc. for the efficient inverse design of metasurfaces. As future work, the generalization capability of the model can be extended to broader design spaces.

ACKNOWLEDGMENT

We express our gratitude to CSIR-4PI for providing access to HPC facility for training the model. We also express our gratitude to Defence Materials and Stores Research and Development Establishment (DMSRDE), DRDO, Kanpur for supporting measurements. The photographs of measurement set-ups included in the paper have been provided by DMSRDE..

REFERENCES

[1] V. Joy, A. Dileep, P. V. Abhilash, R. U. Nair, and H. Singh, "Metasurfaces for stealth applications: A comprehensive review," *Journal of Electronic Materials*, vol. 50, no. 6, pp. 3129–3148, 2021.

[2] H. Hao, S. Du, and T. Zhang, "Radar cross section reduction of metasurface using optimization algorithm," *Progress in Electromagnetics Research M*, vol. 81, pp. 97–106, 2019.

[3] B. Lin, W. Huang, Y. Yang, J. Guo, L. Lv, and J. Zhang, "An ultra-wideband polarization conversion metasurface for RCS reduction," *Journal of Electromagnetic Waves and Applications*, vol. 36, no. 5, pp. 597–606, 2021, doi: 10.1080/09205071.2021.1977185.

[4] B. Lin, W. Huang, J. Guo, et al., "An absorptive coding metasurface for ultra-wideband radar cross-section reduction," *Scientific Reports*, vol. 14, Art. no. 12397, 2024, doi: 10.1038/s41598-024-63260-z.

[5] K. K. Indhu, A. P. Abhilash, R. Anilkumar, et al., "Simplistic metasurface design approach for incident angle and polarization insensitive RCS reduction," *Scientific Reports*, vol. 14, Art. no. 21964, 2024, doi: 10.1038/s41598-024-72509-6.

[6] Q. Guo, Q. Peng, M. Qu, J. Su, and Z. Li, "Optical transparent metasurface for dual-band Wi-Fi shielding," *Optics Express*, vol. 30, no. 5, pp. 7793–7805, 2022.

[7] Y. Yang, W. Li, K. N. Salama, and A. Shamim, "Polarization insensitive and transparent frequency selective surface for dual band GSM shielding," *IEEE Transactions on Antennas and Propagation*, vol. 69, no. 5, pp. 2779–2789, 2021.

[8] M. M. Masud, B. Ijaz, A. Iftikhar, M. N. Rafiq, and B. D. Braaten, "A reconfigurable dual-band metasurface for EMI shielding of specific electromagnetic wave components," in *Proc. IEEE Int. Symp. Electromagnetic Compatibility*, Denver, CO, USA, 2013, pp. 640–644, doi: 10.1109/ISEMC.2013.6670490.

[9] M. Selvaraj, R. Vijay, and R. Anbazhagan, "Reflective metasurface for 5G & beyond wireless communications," *Scientific Reports*, vol. 15, Art. no. 126, 2025, doi: 10.1038/s41598-024-84523-9.

[10] Q. Xiong, Z. Zhang, C. Huang, M. Pu, J. Luo, Y. Guo, J. Ye, W. Pan, X. Ma, L. Yan, and X. Luo, "Amplitude-phase independently encoding space-division multiplexed wireless communication using beamforming reconfigurable metasurfaces," *Advanced Optical Materials*, vol. 12, no. 28, 2024, doi: 10.1002/adom.202401181.

[11] L. Li, H. Zhao, C. Liu, et al., "Intelligent metasurfaces: control, communication and computing," *eLight*, vol. 2, Art. no. 7, 2022, doi: 10.1186/s43593-022-00013-3.

[12] X. Ni, A. Kildishev, and V. Shalaev, "Metasurface holograms for visible light," *Nature Communications*, vol. 4, Art. no. 2807, 2013, doi: 10.1038/ncomms3807.

[13] Y. Hu, X. Luo, Y. Chen, et al., "3D-Integrated metasurfaces for full-colour holography," *Light: Science & Applications*, vol. 8, Art. no. 86, 2019, doi: 10.1038/s41377-019-0198-y.

[14] D. Liu, Y. Tan, E. Khoram, and Z. Yu, "Training deep neural networks for the inverse design of nanophotonic structures," *ACS Photonics*, vol. 5, no. 4, pp. 1365–1369, Apr. 2018.

[15] S. Koziel and S. Ogurtsov, "Simulation-driven design in microwave engineering: Methods," in *Computational Optimization, Methods and Algorithms*, S. Koziel and X.-S. Yang, Eds. Berlin, Heidelberg: Springer, 2011, vol. 356, *Studies in Computational Intelligence*, pp. 231–261, doi: 10.1007/978-3-642-20859-1_8.

[16] J. Hou, H. Lin, W. Xu, X. Shi, R. Tang, L. Chen, and Y. Tian, "Customized inverse design of metamaterial absorber based on target-driven deep learning method," *IEEE Access*, vol. 8, pp. 211849–211859, 2020.

[17] Z. Hou, T. Tang, J. Shen, C. Li, and F. Li, "Prediction network of metamaterial with split ring resonator based on deep learning," *Nanoscale Research Letters*, vol. 15, no. 1, p. 79, Apr. 2020

[18] Z. Tao, J. Zhang, J. You, H. Hao, H. Ouyang, Q. Yan, S. Du, Z. Zhao, Q. Yang, X. Zheng, and T. Jiang, "Exploiting deep learning network in optical chirality tuning and manipulation of diffractive chiral metamaterials," *Nanophotonics*, vol. 9, no. 9, pp. 2945–2956, 2020

[19] Z. Tao, J. You, J. Zhang, X. Zheng, H. Liu, and T. Jiang, "Optical circular dichroism engineering in chiral metamaterials utilizing a deep learning network," *Optics Letters*, vol. 45, no. 6, pp. 1403–1406, Mar. 2020, doi: 10.1364/OL.386980.

[20] M. Raissi, P. Perdikaris, and G. E. Karniadakis, "Physics-informed neural networks: A deep learning framework for solving forward and inverse problems involving nonlinear partial differential equations," *Journal of Computational Physics*, vol. 378, pp. 686–707, 2019.

[21] L. Lu, X. Meng, Z. Mao, and G. E. Karniadakis, "DeepXDE: A deep learning library for solving differential equations," *SIAM Review*, vol. 63, no. 1, pp. 208–228, 2021.

[22] O. Hennigh, S. Narasimhan, M. A. Nabian, A. Subramaniam, K. Tangsali, Z. Fang, M. Rietmann, W. Byeon, and S. Choudhry, "NVIDIA SimNet™: An AI-accelerated multi-physics simulation framework," in *Proc. Computational Science – ICCS 2021: 21st Int. Conf.*, Krakow, Poland, Jun. 16–18, 2021, Part V, Springer, pp. 447–461.

[23] Z. Peng, B. Yang, L. Liu, and Y. Xu, "Rapid surrogate modeling of magnetotelluric in the frequency domain using physics-driven deep neural networks," *Computers & Geosciences*, vol. 176, Art. no. 105360, 2023, doi: 10.1016/j.cageo.2023.105360.

[24] O. Noakoasteen, C. Christodoulou, Z. Peng, and S. K. Goudos, "Physics-informed surrogates for electromagnetic dynamics using Transformers and graph neural networks," *IET Microwaves, Antennas & Propagation*, vol. 18, no. 7, pp. 505–515, Feb. 2024, doi: 10.1049/mia2.12463.

[25] G. Jing, P. Wang, H. Wu, J. Ren, Z. Xie, J. Liu, H. Ye, Y. Li, D. Fan, and S. Chen, "Neural network-based surrogate model for inverse design of metasurfaces," *Photonics Research*, vol. 10, pp. 1462–1471, 2022.

[26] A. Mall, A. Patil, A. Sethi, and A. Kumar, "A cyclical deep learning based framework for simultaneous inverse and forward design of nanophotonic metasurfaces," *Scientific Reports*, vol. 10, no. 1, p. 19427, Nov. 2020.

[27] S. An, B. Zheng, M. Y. Shalaginov, H. Tang, H. Li, L. Zhou, J. Ding, A. M. Agarwal, C. Rivero-Baleine, M. Kang, K. A. Richardson, T. Gu, J. Hu, C. Fowler, and H. Zhang, "Deep learning modeling approach for metasurfaces with high degrees of freedom," *Optics Express*, vol. 28, no. 21, pp. 31932–31942, Oct. 2020.

[28] H. P. Wang, D. M. Cao, X. Y. Pang, X. H. Zhang, S. Y. Wang, W. Y. Hou, C. C. Nie, and Y. B. Li, "Inverse design of metasurfaces with customized transmission characteristics of frequency band based on generative adversarial networks," *Optics Express*, vol. 31, no. 23, pp. 37763–37777, Nov. 2023.

[29] M. Y. Li, E. Grant, and T. L. Griffiths, "Gaussian process surrogate models for neural networks," in *Proc. Uncertainty in Artificial Intelligence (UAI)*, PMLR, 2023.

[30] M. M. R. Elsawy, M. Binois, R. Duvigneau, S. Lanteri, and P. Genevet, "Optimization of metasurfaces under geometrical uncertainty using statistical learning," *Optics Express*, vol. 29, pp. 29887–29898, 2021.

[31] J. P. Jacobs and S. Koziel, "Two-stage Gaussian process modeling of microwave structures for design optimization," in *Simulation-Driven Modeling and Optimization*, S. Koziel, L. Leifsson, and X.-S. Yang, Eds. Cham, Switzerland: Springer, 2016, vol. 153, *Springer Proceedings in Mathematics & Statistics*, pp. 109–127, doi: 10.1007/978-3-319-27517-8_7.

[32] D. I. L. de Villiers, I. Couckuyt and T. Dhaene, "Multi-objective optimization of reflector antennas using kriging and probability of improvement," 2017 IEEE International Symposium on Antennas and Propagation & USNC/URSI National Radio Science Meeting, San Diego, CA, USA, 2017, pp. 985-986, doi:

[33] T. Hastie, R. Tibshirani, and J. Friedman, The elements of statistical learning: data mining, inference, and prediction, 2nd ed. New York, NY, USA: Springer, 2009.

[34] B. A. Munk, *Finite Antenna Arrays and FSS*. Hoboken, NJ, USA: Wiley, 2003.

[35] CST Studio Suite®, CST AG, Germany, www.cst.com.